\documentclass[a4paper,twoside]{article}

\usepackage{epsfig}
\usepackage{subfigure}
\usepackage{calc}
\usepackage{amssymb}
\usepackage{amstext}
\usepackage{amsmath}
\usepackage{amsthm}
\usepackage{multicol}
\usepackage{pslatex}
\usepackage{apalike}
\usepackage{SCITEPRESS}     
\usepackage{multirow}
\usepackage{threeparttable}
\usepackage{float}

\begin{document}

\title{Features Extraction Based on an Origami Representation of 3D Landmarks}
\author{\authorname{Juan Manuel Fernandez Montenegro\sup{1}, Mahdi Maktab Dar Oghaz\sup{1}, Athanasios~Gkelias\sup{2}, Georgios~Tzimiropoulos\sup{3}, and Vasileios~Argyriou\sup{1}}
\affiliation{\sup{1}Kingston University London, \sup{2}Imperial College London, \sup{3}University of Nottingham}
\email{\{J.Fernandez, M.Maktabdaroghaz, Vasileios.Argyriou\}@kingston.ac.uk, A.Gkelias@imperial.ac.uk, yorgos.tzimiropoulos@nottingham.ac.uk}
}
\abstract{Feature extraction analysis has been widely investigated during the last decades in computer vision community due to the large range of possible applications. Significant work has been done in order to improve the performance of the emotion detection methods. Classification algorithms have been refined, novel preprocessing techniques have been applied and novel representations from images and videos have been introduced. In this paper, we propose a preprocessing method and a novel facial landmarks' representation aiming to improve the facial emotion detection accuracy. We apply our novel methodology on the extended Cohn-Kanade (CK+) dataset and other datasets for affect classification based on Action Units (AU). The performance evaluation demonstrates an improvement on facial emotion classification (accuracy and F1 score) that indicates the superiority of the proposed methodology.}

\onecolumn \maketitle

\section{\uppercase{Introduction}}

Face analysis and particularly the study of human affective behaviour has been part of many disciplines for several years such as computer science, neuroscience or psychology \cite{Zeng2009}. The accurate automated detection of human affect can benefit areas such as Human Computer Interaction (HCI), mother-infant interaction, market research, psychiatric disorders or dementia detection and monitoring. Automatic emotion recognition approaches are focused on the variety of human interaction capabilities and biological data. For example, the study of speech and other acoustic cues in~\cite{Weninger2015,Chowdhuri2016}, body movements in~\cite{Stock2015}, electroencephalogram (EEG) in~\cite{Lokannavar2015}, facial expressions~\cite{FERA2017,Song2015,Baltru2016} or combinations of previous ones such as speech and facial expressions in~\cite{Nicolaou2011} or EEG and facial expressions in~\cite{Soleymani2016}.

One of the most popular facial emotion model is the Facial Action Coding System (FACS)~\cite{Ekman1978}. It describes facial human emotions such as happiness, sadness, surprise, fear, anger or disgust; where each of these emotions is represented as a combination of Action Units (AUs). Other approaches abandon the path of specific emotions recognition and focus on emotions' dimensions, measuring their valence, arousal and intensity~\cite{Nicolaou2011,Nicolle2012,Zhao2009}, or pleasantness-unpleasantness, attention-rejection and sleep-tension dimensions in the three dimension Schlosberg Model~\cite{Izard2013}. When it comes to the computational affect analysis, the methods for facial emotion recognition can be classified according to the approaches used during the recognition stages: registration, features selection, dimensionality reduction or classification/recognition~\cite{Alpher2015,Bettadapura2012,Sariyanidi2013,Chu2017,Gudi2015,Yan2017}.

Most of the state of the art approaches for facial emotion recognition use posed datasets for training and testing such as CK~\cite{Kanade2000} and MMI~\cite{Pantic2005}. These datasets provide data on non-naturalistic conditions regarding illumination or nature of expression. In order to have more realistic data, non-posed datasets were created such as SEMAINE~\cite{McKeown2012}, MAHNOB-HCI~\cite{Soleymani2012}, SEMdb \cite{Montenegro2016,FERNANDEZMONTENEGRO201742}, DECAF~\cite{Abadia2015}, CASME II~\cite{Yan2014}, or CK+~\cite{Lucey2010}. On the other hand, some applications do require a controlled environment, therefore, posed datasets can be more suitable to certain applications.

A crease pattern is the underline blueprint for an origami figure. The universal molecule is a crease pattern constructed by the reduction of polygons until they are reduced to a point or a line. Lang~\cite{Lang1996} presented a computational method to produce crease patterns with a simple uncut square of paper that describes all the folds necessary to create an origami figure. Lang's algorithm proved that it is possible to create a crease pattern from a shadow tree projection of a 3D model. This shadow tree projection is like a dot and stick molecular model where the joints and extremes of a 3D model are represented as dots and the connections as lines. 

The use of Eulerian magnification~\cite{Wu2012,Wadhwa2016} has been proved to increase the classification results for facial emotion analysis. The work presented in~\cite{Park2015} uses Eulerian magnification on a spatio-temporal approach that recognises five emotions using a SVM classifier reaching a 70\% recognition rate on CASME II dataset. The authors in~\cite{LeNgo2016} obtained an improvement of $0.12$ in the $F1$ score using a similar approach.

Amongst the most common classifiers for facial emotion analysis, SVM, boosting techniques and artificial neural networks (e.g. DNNs) are the most used~\cite{Lokannavar15,Yi2013,FERA2017}. The input to these classifiers are a series of features extracted from the available data that will provide distinctive information of the emotions such as facial landmarks, histogram of gradients (HOG) or SIFT descriptors~\cite{Corneanu2016}. The purpose of this work is to introduce novel representations and preprocessing methods for face analysis and specifically facial emotion classification and to demonstrate the improvement of the classification results. The proposed preprocessing methodology uses Eulerian magnification in order to enhance facial movements as presented in~\cite{Wadhwa2016}, which provides a more pronounced representation of facial expressions. The proposed representation is based on Lang's Universal Molecule Algorithm~\cite{Bowers2015} resulting in a crease pattern of the facial landmarks. In summary, the main contributions are:
a) We suggested a motion magnification approach as a preprocessing stage aiming to enhance facial micro-movements improving the overall emotion classification accuracy and performance. b) We developed a new Origami based methodology to generate novel descriptors and representations of facial landmarks. c) We performed rigorous analysis and demonstrated that the addition of the proposed descriptors improve the classification results of state-of-the-art methods based on Action Units for face analysis and basic emotion recognition.

The remainder of this paper is organized as follows: Section 2 presents the proposed methodology and in section 3 details on the evaluation process and the obtained results are provided. Section 4 gives some conclusion remarks.

\section{\uppercase{Proposed Methodology}}

\begin{figure}
\centering
  \includegraphics[width=0.9\linewidth]{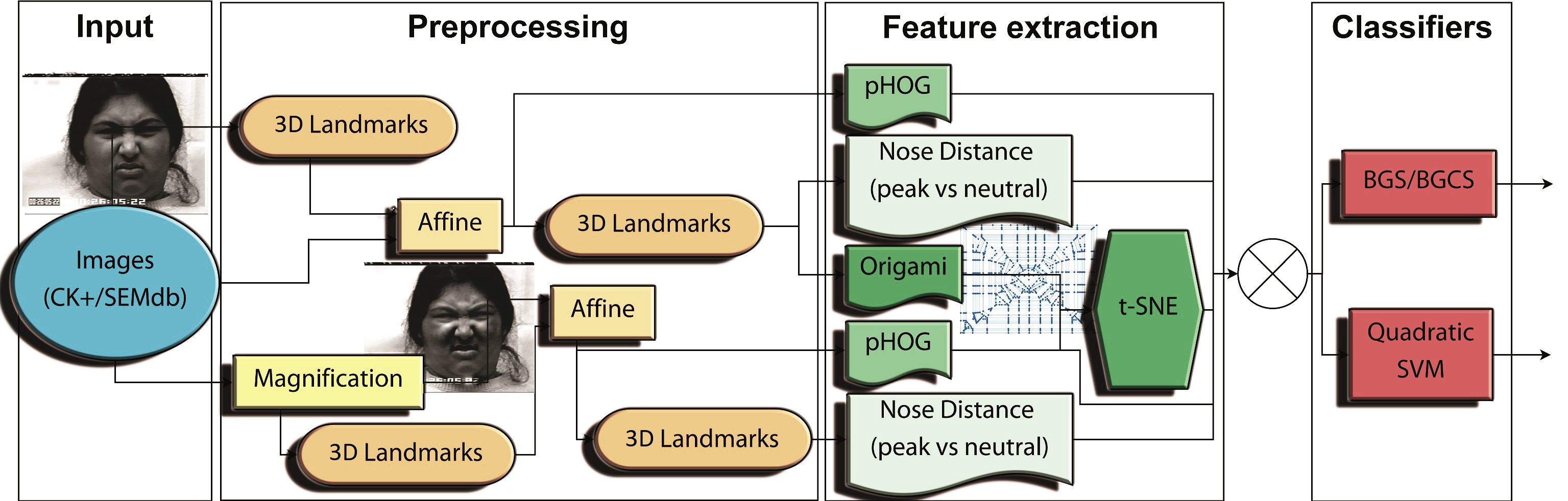}\\
  \caption{The diagram of the proposed methodology visualising the steps from input and preprocessing, to face representation and classification for all the suggested variations considering the classic feature based machine learning solutions.}
  \label{fig:diagram}
\end{figure}

Our proposed methodology comprises a data preprocessing, a face representation and a classification phase. These phases are applied both for feature based learning methods and deep architectures. Regarding the feature based methods (see figure~\ref{fig:diagram}), in the preprocessing phase, facial landmarks are extracted with or without Eulerian Video Magnification and an affine transform is applied for landmark and face alignment. In order to obtain subpixel alignment accuracy the locations are refined using frequency domain registration methods, \cite{1224852,1408930,5508420}. In the face representation phase and the feature based learning approaches, a pyramid Histogram of Gradients (pHOG) descriptor is utilised. After the affine transformation new facial landmarks are extracted and they are used as input both to the Lang's Universal Molecule algorithm to extract the novel origami descriptors and to the facial displacement descriptor (i.e., the Euclidean distance between the nose location in a neutral and a `peak' frame representative of the expression). Finally, in the classification phase, the pHOG features are used as input to a Bayesian Compressed Sensing (BCS) and a Bayesian Group-Sparse Compressed Sensing (BGCS) classifiers, while the origami and nose-distance displacement descriptors are provided as inputs to a Quadratic Support Vector Machine (SVM) classifier. The t-Distributed Stochastic Neighbor Embedding (tSNE) dimensionality reduction technique is applied to the origami descriptor before the SVM classifier.

Regarding, the deep neural networks and considering an architecture based on the work on Attentional Recurrent Relational Network-LSTM (ARRN-LSTM) or the Spatial-Temporal Graph Convolutional Networks (ST-GCN)~\cite{Li2018,Yan2018}, while the preprocessing stage remains the same. The input images are processed to extract facial landmarks, and then aligned. Furthermore, the process is applied twice and the second time for the motion magnified images. During the face representation the proposed origami representation is applied on the extracted landmarks and then the obtained pattern is imported to the above networks aiming to model simultaneously both spatial and temporal information. The utilised methodology is summarized in Figure~\ref{fig:diagram} and the whole process is described in detail in the following subsections.

\subsection{Preprocessing Phase}

The two main techniques used in the data preprocessing phase (apart from the facial landmarks extraction) are the affine transform for landmark alignment and the Eulerian Video Magnification (EVM). In the preprocessing phase a set of output sequences are generated based on a combination of different techniques.

\textbf{SEC-1}: The first output comprises affine transformation of the original sequence of images which are then rescaled to 120$\times$120 pixels and converted to gray-scale. This represents the input of the pHOG descriptor.

\textbf{SEC-2}: The second output comprises Eulerian magnified images which are then affine transformed, rescaled to 120$\times$120 pixels and converted to gray scale. This represents again the input of the pHOG descriptor.

\textbf{SEC-3}: In the third output, the approaches proposed in~\cite{Kumar2018,Baltru2016,ARGYRIOU20091} are used before and after the affine transformation to reconstruct the face and obtain the 3D facial landmarks. This is provided as input to the facial displacement and the proposed origami descriptor.

\textbf{SEC-4}: In the fourth output, the facial landmarks are obtained from the Eulerian motion magnified images, the affine transform is applied and the facial landmarks are estimated again. This is given as input to the facial feature displacement descriptor.

The same process is considered for the deep ARRN-LSTM architecture with input the obtained 3D facial landmarks extracted by the proposed origami transformation.

\subsection{Feature Extraction Phase}

Three schemes have been used in this phase for the classic feature based learning approaches, (1) the pyramid Histogram of Gradients (pHOG), (2) a facial feature displacement descriptor, and (3) the proposed origami descriptor. These schemes are analysed below.

\textbf{pHOG features extraction}: The magnified affine transformed sequence (i.e., SEC-2) is provided as input to the pHOG descriptor. More specifically, eight bins on three pyramid levels of the pHOG are applied in order to obtain a row of $h_m$ features per sequence. For comparison purposes, the same process is applied to the unmagnified sequence (i.e., SEC-1) and a row of $h$ features per sequence is obtained respectively.

\textbf{Facial feature displacement}: The magnified facial landmarks (i.e., fourth sequence output of the preprocessing phase) are normalised according to a fiducial face point (i.e., nose) to account for head motion in the video stream. In other words, the nose is used as a reference point, such that the position of all the facial landmarks are independent of the location of the subject's head in the images. If $L_i=[L_{i_x}\quad L_{i_y}]$ are the original image coordinates of the $i$-th landmark, and $L_n=[L_{n_x}\quad L_{n_y}]$ the nose landmark coordinates, the normalized coordinates are given by $l_i=[L_{i_x}-L_{n_x}\quad L_{i_y}-L_{n_y}]$.

The facial displacement features are the distances between each facial landmark in neutral pose and the corresponding ones in the `peak' frame that represents the corresponding expression. The displacement of the $i$-th landmark (i.e., $i$-the vector element) is calculated using the Euclidean distance
\begin{equation}
d(l_i^{(p)},l_i^{(n)})=\sqrt{(l_{i_x}^{(p)}-l_{i_x}^{(n)})^2- (l_{i_y}^{(p)}-l_{i_y}^{(n)})^2}
\end{equation}
between its normalised position in neutral frame ($l_i^{(n)}$) and the `peak' frame ($l_i^{(p)}$). In the remaining of this paper, we will be referring to these features as distance to nose (neutral vs peak) features (DTNnp). The output of the DTNnp descriptor is a $d_m$ long row vector per sequence. For comparison purposes, the same process is applied to unmagnified facial landmark sequences (i.e., the third sequence of the preprocessing phase) and a row of $d$ features per sequence is obtained accordingly.

\subsubsection{Origami based 3D face Representation}
The origami representation is created from the normalised facial 3D landmarks (i.e., SEC-3). The descriptor is using $o$ facial landmarks in order to create an undirected graph of $n$ nodes and $e$ edges representing the facial crease pattern. The $n$ nodes contain the information of the $x$ and $y$ landmark coordinates, while the $e$ edges contain the IDs of the two nodes connected by the corresponding edge (which represent the nodes/landmarks relationships). The facial crease pattern creation process is divided into three main steps: shadow tree, Lang's polygon and shrinking.

\begin{figure}
\centering
  \includegraphics[scale = 0.2]{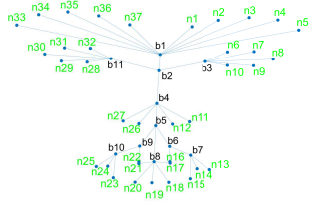}
  \caption{Shadow tree. The facial landmarks are linked creating a symmetric tree.}
  \label{fig:shadowTree}
\end{figure}

The first step implies the extraction of the flap projection of the face (shadow tree), from the facial landmarks. This shadow tree or metric tree $(T,d)$ is composed of leaves (external nodes) $N_ex = {n_1,...,n_p}$, internal nodes $N_in = {b_1,...,b_q}$, edges $E$ and distances $d$ to the edges. This distance is the Euclidean distance between connected landmarks (nodes) through an edge in the tree. It is just a distance measured between each landmark that is going to be used during the Lang's Polygon creation and during the shrinking step. The shadow tree is created as a linked version of 2D facial landmarks of the eyebrows, eyes, nose and mouth (see Figure~\ref{fig:shadowTree}).

\begin{figure}
\centering
  \includegraphics[scale = 0.2]{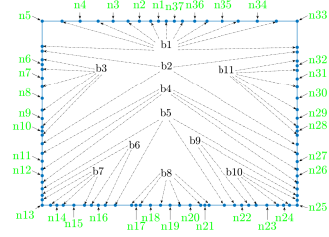}
  \caption{Based on Lang's polygon rules, a rectangle-shaped initial convex doubling cycle polygon was created from the shadow tree in order to start with the same initial polygon shape for any face. The dashed arrows point the correspondent internal nodes $b_i$ and the straight arrows point the leave nodes $n_i$.}
  \label{fig:langPolygonSquare}
\end{figure}

During the second step a convex doubling cycle polygon (Lang's polygon) $L_p$, is created from the shadow tree $(T,d)$, based on the double cycling polygon $f$ creation algorithm. According to this process we are walking from one leaf node to the next one in the tree until reaching the initial node. Therefore, the path to go from $n_1$ to $n_2$ includes the path from $n_1$ to $b_1$ and from $b_1$ to $n_2$; the path from $n_2$ to $n_3$ also requires to pass through $b_1$; and the path from $n_5$ to $n_6$ goes through $b_1$, $b_2$ and $b_3$. In order to guaranty the resultant polygon to be convex, we shaped it as a rectangle (see Figure~\ref{fig:langPolygonSquare}), where the top side contains the landmarks of the eyebrows, the sides are formed by the eyes and nose, and the mouth landmarks are at the bottom. This Lang's polygon represents the area of the face that is going to be folded.

The obtained convex polygonal region has to satisfy the following condition: the distance between the leaf nodes $n_i$ and $n_j$ in the polygon ($d_P$) should be equal or greater than the distance of those leaf nodes in the shadow tree ($d_T$). This requirement is mainly due to the origami properties, so since a shadow tree come from a folded piece of paper (face), once it is unfolded to see the crease pattern, the distances on the unfolded paper $d_P(n_i,n_j)$ are going to be always larger or equal to the distances in the shadow tree $d_T(n_i,n_j)$.
\begin{equation}\label{eq=oriCondition}
   d_P(n_i,n_j)\geq d_T(n_i,n_j)
\end{equation}

\begin{figure}
\centering
  \includegraphics[scale = 0.2]{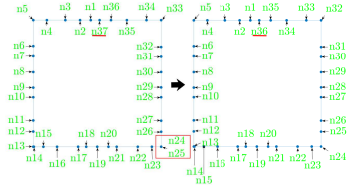}
  \caption{Contraction event. In this case the distance between nodes 24 and 25 is 0 so one of the nodes is eliminated. Therefore the number of leaf nodes is reduced to 36. Nodes 13 and 14 are close but not close enough to trigger a contraction event.}
  \label{fig:contraction}
\end{figure}

The third step corresponds to the shrinking process of this polygon. All the edges are moving simultaneously towards the centre of the polygon at constant speed until one of the following two events occur: contraction or splitting. The contraction event happens when the points collide in the same position (see Eq.~\ref{eq=contractCond}). In this case, only one point is kept and the shrinking process continues (see Figure~\ref{fig:contraction}).
\begin{equation}\label{eq=contractCond}
   d_P(n_i,n_j)\leq th, \quad \textrm{then} \quad n_i = n_j
\end{equation}
where $j=i+1$ and $th$ is a positive number $\approx0$.

\begin{figure}
\centering
  \includegraphics[scale = 0.2]{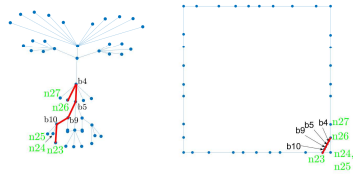}
  \caption{Splitting event. When the distance between two non-consecutive points in Lang's polygon (nodes 23 and 26) is the same as the distance of those two points in the shadow tree, a new edge is created. The intermediate nodes in the tree are also located in the new edge (nodes b4, b5, b9, b10).}
  \label{fig:splitting0}
\end{figure}

The splitting event occurs when the distance between two non-consecutive points is equal to their shadow tree distance (see Figure~\ref{fig:splitting0}).
\begin{equation}\label{eq=splitCond}
   d_P(n_i,n_k)\leq d_T(n_i,n_k) + th
\end{equation}
where $k\geq i+1$ and $th$ is a positive number $\approx0$. As a consequence of this event a new edge is generated between these points creating two new sub-polygons. The shrinking process continues on each polygon separately.
 

\begin{figure}
\centering
  \includegraphics[scale = 0.18]{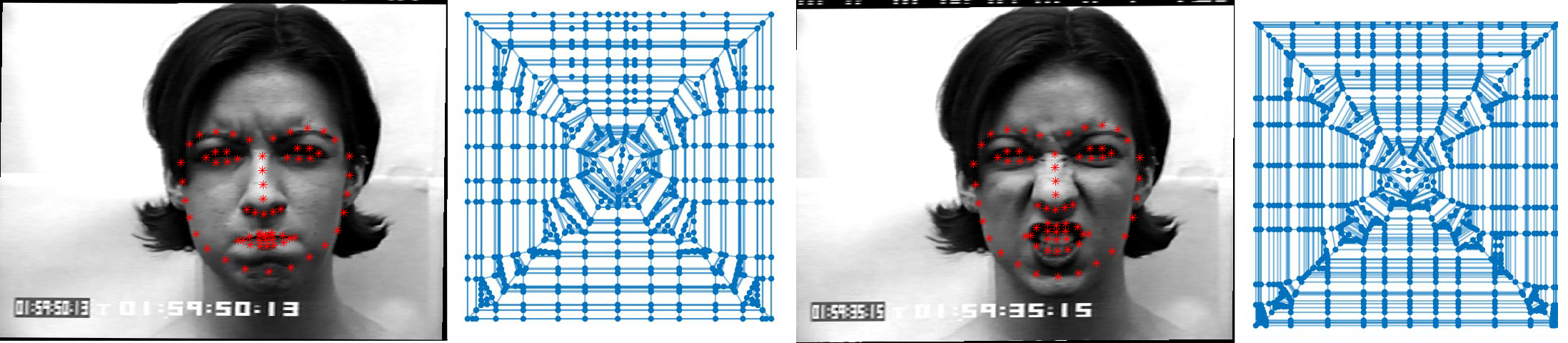}
  \caption{This image shows the crease pattern from the proposed algorithm applied to our rectangular convex doubling cycle polygon produced from the facial landmarks.}
  \label{fig:creasePattern}
\end{figure}

Finally, the process will end when all the points converge, creating a crease pattern, $C = g(f({T,d}))$, where $(T,d)$ is the shadow tree, $f(T,d)$ is the Lang polygon, $C$ is the crease pattern and $g$ and $f$ are the processes to create the double cycle polygon and shrinking functions, respectively. Due to the structure of our initial Lang's polygon, the final crease patterns will have a structure similar to the one shown in Figure~\ref{fig:creasePattern}.

The crease pattern is stored as an undirected graph containing the coordinates of each node and the required information for the edges (i.e. IDs of linked nodes). Due to the fact that $x$ and $y$ coordinates and the linked nodes are treated as vectorial features, these can be represented more precisely by a complex or hyper-complex representation, as shown in~\cite{Adali2011}. A vector can be decomposed into linearly independent components, in the sense that they can be combined linearly to reconstruct the original vector. However, depending on the phenomenon that changes the vector, correlation between the components may exist from a statistical point of view (i.e. two uncorrelated variables are linearly independent but two linearly independent variables are not always uncorrelated). If they are independent our proposed descriptor does not provide any significant advantage, but if there is correlation this is considered. In most of the cases during the feature extraction process complex or hyper-complex features are generated but decomposed to be computed by a classifier~\cite{Adali2011,Li2011}. In our case, the coordinates of the nodes are correlated and also the node IDs that represent the edges. Therefore, the nodes and edges are represented as show in~\ref{eq=nodes} and~\ref{eq=edges}.
\begin{equation}\label{eq=nodes}
   n = {n_i}_x + i· {n_i}_y
\end{equation}
\begin{equation}\label{eq=edges}
   e = {{e_i}\_nodeID_1 + i·{e_i}\_nodeID_2}
\end{equation}
where ${n_i}_x$ and ${n_i}_y$ is the coordinate $x$ and $y$ of the $i_th$ leave node; and ${e_i}\_nodeID_1$ and ${e_i}\_nodeID_2$ are the identifiers of the nodes linked by the $ith$ edge.

\subsection{Classification Phase}
\subsubsection{Feature based classification approaches}
Two state-of-the-art methods have been used in the third and last phase of the proposed emotion classification approach. The first method is based on a Bayesian Compressed Sensing (BCS) classifier, including its improved version for Action Units detection, and a Bayesian Group-Sparse Compressed Sensing (BGCS) classifier, similar to the one presented in~\cite{Song2015}. The second method is similar to~\cite{Michel2003} and is based on a quadratic SVM classifier which has been successfully applied for emotion classification in~\cite{Buciu2003}.

\section{\uppercase{Results}}

Data from two different datasets (CK+ and SEMdb) is used to validate the classification performance of the methods by calculating the classification accuracy and the F1 score. The CK+ dataset contains 593 sequences of images from 123 subjects. Each sequence starts with a neutral face and ends with the peak stage of an emotion. The CK+ contains AU labels for all of them but basic emotion labels only for 327. The SEMdb contains 810 recordings from 9 subjects. The start of each recording is considered as a neutral face and the peak frame is the one whose landmarks vary most from the respective landmarks in the neutral face. SEMdb contains labels for 4 classes related to autobiographical memories. These autobiographical memory classes are represented by spontaneous facial micro-expressions triggered by the observation of 4 different stimulations related to distant and recent autobiographical memories.

The next paragraphs explain the obtained results ordered by the objective classes (AUs, 7 basic emotions or 4 autobiographical emotions). The experiments are compared with results obtained using two state of the art methods: Bayesian Group-Sparse Compressed Sensing~\cite{Song2015} and landmarks to nose~\cite{Michel2003}. Song et al. method compares two classification algorithms, i.e., the Bayesian Compressed Sensing (BCS) and their proposed improvement Bayesian Group-Sparse Compressed Sensing (BGCS). Both classifiers are used to detect 24 Action Units within the CK+ dataset using pHOG features. Landmarks to nose methods consist on using the landmarks distance to nose difference between the peak and neutral frame as input of an SVM classifier to classify 7 basic emotions in ~\cite{Michel2003} or 4 autobiographical emotions in \cite{Montenegro2016}. Regarding the DNN architectures the comparative study is performed with the original method proposed by Yan in~\cite{Yan2018} considering both the origami graph in a Graph Convolutional network with and without Eulerian Video Magnification.

\begin{figure}
\centering
  \includegraphics[scale = 0.2]{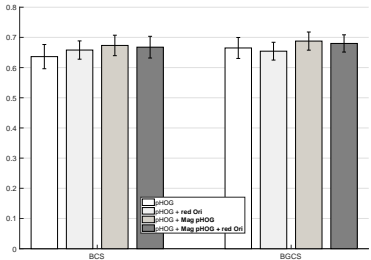}
  \caption{F1 score results of the 24 AUs classification using the state of the art methods BCS and BGCS and the new results when the novel features are added.}
  \label{fig:barsBCSBGCSresults}
\end{figure}

The 24 AUs detection experiment involved the BCS and the BGCS classifiers and CK+ database. The input features utilised included the state of the art ones and them combined with our proposed ones. Therefore, for the AUs experiment the pHOG were tested independently and combined with the proposed magnified pHOG and origami features; and identically with the distance to nose features. The result of the different combinations are shown in Figure~\ref{fig:barsBCSBGCSresults}, Table~\ref{tab:BCSCK+} and Table~\ref{tab:BCScombCK}. They show that the contribution of the new descriptors improves the F1 score and the overall accuracy.

\begin{table}[!t]
  \centering
  \caption{Mean of k-fold F1 score and Accuracy values. 24 AU from CK+ dataset are classified using 4 combinations of features and BCS and BGCS classifier}
  \setlength\tabcolsep{2pt}
    \begin{tabular}{|r|r|r|r|r|r|}
    \hline
     \multirow{ 4}{*}{Method} & \multirow{ 4}{*}{Meas.} &  \multicolumn{ 4}{c|}{Features}\\  \cline{3-6}
     &  & \multicolumn{1}{c|}{pHOG} & \multicolumn{1}{c|}{pHOG}  & \multicolumn{1}{c|}{pHOG}   & \multicolumn{1}{c|}{pHOG} \\
     &  &  &  \textbf{Ori} &  \textbf{Mag pHOG} &  \multicolumn{1}{c|}{\textbf{Mag pHOG}}\\
     &  &  &   &   &  \multicolumn{1}{c|}{\textbf{red Ori}}\\
    \hline
    \multirow{ 2}{*}{BCS}  & F1   & 0.636 & 0.658 & \textbf{ 0.673} & 0.667\\
      & ACC  & 0.901 & 0.909 & 0.909 & \textbf{0.912}\\ \hline
    \multirow{ 2}{*}{BGCS} & F1   & 0.664 & 0.654 & \textbf{ 0.687} & 0.679\\
     & ACC  & 0.906 & 0.908 & \textbf{ 0.913} & 0.91 \\ \hline
   \end{tabular}%
  \label{tab:BCSCK+}%
\end{table}%

\begin{table}[!t]
  \centering
  \caption{Mean of k-fold F1 score and Accuracy values. 24 Action Units from CK+ dataset are classified. A combination of the DTNnp~\cite{Montenegro2016}, the pHOG~\cite{Song2015} extracted from the magnified version of the data and the novel origami features are used as input to the BCS/BGCS classifiers.}
  \setlength\tabcolsep{2pt}
    \begin{tabular}{|r|r|r|r|r|}
    \hline
    \multirow{ 4}{*}{Method} & \multirow{ 4}{*}{Meas.} &  \multicolumn{ 3}{c|}{Features}\\  \cline{3-5}
      &  &  \multicolumn{1}{c|}{DTNnp} &  \multicolumn{1}{c|}{DTNnp} &  \multicolumn{1}{c|}{DTNnp} \\
    &  &  & \textbf{Mag pHOG} & \multicolumn{1}{c|}{\textbf{Mag pHOG}}\\
    &  &  &  &\multicolumn{1}{c|}{\textbf{Ori}}\\
    \hline
    \multirow{ 2}{*}{BCS}  & F1   & 0.407 & \textbf{0.671} & 0.663\\
      & ACC  & 0.873 & 0.913 & \textbf{0.914}\\ \hline
    \multirow{ 2}{*}{BGCS}  & F1   & 0.555 & \textbf{0.69} & 0.682\\
      & ACC  & 0.893 & \textbf{0.915} & 0.914\\
    \hline

    \end{tabular}%
  \label{tab:BCScombCK}%
\end{table}%

The second experiment's objective was the classification of 7 basic emotions using a quadratic SVM classifier, the distance to nose features combined with the proposed ones and the CK+ database. The results shown in Table~\ref{tab:SVMcombCK} the quadratic SVM ones. The combination of the features did not provide any noticeable boost in the accuracy. The increment in the F1 score is not big enough to be taken into account. Similar results were obtained for the VGG-S architecture with an increment in accuracy and the F1 score.

\begin{table}[!t]
  \centering
  \caption{Mean of k-fold F1 score and Accuracy values. 7 emotion classes from CK+ dataset are classified. A combination of the DTNnp~\cite{Montenegro2016}, the pHOG~\cite{Song2015} extracted from the magnified version of the data and the novel origami features are used as input to the quadratic SVM classifier. The final rows demonstrate the results obtained using the VVG-S architectures.}
  \setlength\tabcolsep{2pt}
    \begin{tabular}{|r|r|r|r|r|}
    \hline
    \multirow{ 4}{*}{Method} & \multirow{ 4}{*}{Meas.} &  \multicolumn{ 3}{c|}{Features}\\  \cline{3-5}
      &  &  \multicolumn{1}{c|}{DTNnp} &  \multicolumn{1}{c|}{DTNnp} &  \multicolumn{1}{c|}{DTNnp} \\
    &  &  &  \textbf{Mag pHOG} &  \multicolumn{1}{c|}{\textbf{Mag pHOG}}\\
    &  &  &  &  \multicolumn{1}{c|}{\textbf{Ori}}\\
    \hline
    \multirow{ 2}{*}{qSVM}  & F1   & 0.861 & 0.857 & \textbf{0.865}\\
      & ACC  & \textbf{0.899} & 0.893 & \textbf{0.899}\\
      \hline
    Method & Meas. & Orig & Origami & Origami + Mag\\
    \hline\multirow{ 2}{*}{VGG-S}  & F1   & 0.269 & 0.282 & \textbf{0.284}\\
      & ACC  & 0.24 & 0.253 & \textbf{0.26}\\
    \hline
    \end{tabular}%
  \label{tab:SVMcombCK}%
\end{table}%

\begin{table}[!t]
  \centering
  \caption{The obtained confusion matrix for the 7 emotions present in the CK+ dataset.}
  \setlength\tabcolsep{1pt}
    \begin{tabular}{l|ccccccc}
          & anger & cont & disg & fear & happy & sadn & surpr \\ \hline
    anger &38  &  2  &  3  &  0  &  0  &  2  &  0 \\
    contempt & 1  &  15   &  0   &  0  &   1   &  1  &  0 \\
    disgust & 5  &  0  &  54   &  0   &  0  &   0   &  0 \\
    fear & 0  &  0  &   0   & 19   &  4  &   0   &  2 \\
    happy & 0  &  2  &   0   &  1   & 66  &   0   &  0 \\
    sadness & 3  &  0  &   1   &  1   &  1  &  22   &  0 \\
    surprise & 0  &  2  &   0   &  0   &  1  &   0   & 80 \\
    \end{tabular}%
  \label{tab:confMatrixSVMck}%
\end{table}%

Table~\ref{tab:confMatrixSVMck} shows the confusion matrix of the emotions classified for the best F1 score experiments. This confusion matrix shows that fear is the emotion with less rate of success and surprise and happiness to be the more accurately recognised. The third experiment involved the BCS and quadratic SVM classifiers, using the SEMdb dataset, pHOG features and a combination of pHOG and distance to nose features (both combined with the proposed ones) and detecting the corresponding 4 classes that are related to autobiographical memories. Both classifiers (BCS and SVM) provide improved classification estimates when the combination of all features is used to detect the 4 autobiographical memory classes.

\section{\uppercase{Conclusion}}
We presented the improvement that novel preprocessing techniques and novel representations can provide in the classification of emotions from facial images or videos. Our study proposes the use of Eulerian magnification in the preprocessing stage and an origami algorithm in the feature extraction stage. Our results show that the addition of these techniques can help to increase the overall classification accuracy both for Graph Convolutional Network and feature based methods.

\section*{\uppercase{Acknowledgements}}
This work is co-funded by the EU-H2020 within the MONICA project under grant agreement number 732350. The Titan X Pascal used for this research was donated by NVIDIA

\bibliographystyle{apalike}
{\small
\bibliography{egbib}}

\begin{thebibliography}{}

\bibitem[Abadia et~al., 2015]{Abadia2015}
Abadia, M.~K., Subramanian, R., Kia, S.~M., Avesani, P., Patras, I., and Sebe,
  N. (2015).
\newblock Decaf: Meg-based multimodal database for decoding affective
  physiological responses.
\newblock {\em IEEE Transactions on Affective Computing}, 6(3):209--222.

\bibitem[Adali et~al., 2011]{Adali2011}
Adali, T., Schreier, P., and Scharf, L. (2011).
\newblock Complex-valued signal processing: The proper way to deal with
  impropriety.
\newblock {\em IEEE Transactions on Signal Processing}, 59(11):5101--5125.

\bibitem[Alpher, 2015]{Alpher2015}
Alpher, A. (2015).
\newblock Automatic analysis of facial affect: A survey of registration,
  representation, and recognition.
\newblock {\em IEEE transactions on pattern analysis and machine intelligence},
  37(6):1113--1133.

\bibitem[Argyriou, 2011]{5508420}
Argyriou, V. (2011).
\newblock Sub-hexagonal phase correlation for motion estimation.
\newblock {\em IEEE Transactions on Image Processing}, 20(1):110--120.

\bibitem[Argyriou and Petrou, 2009]{ARGYRIOU20091}
Argyriou, V. and Petrou, M. (2009).
\newblock Chapter 1 photometric stereo: An overview.
\newblock In Hawkes, P.~W., editor, {\em Advances in IMAGING AND ELECTRON
  PHYSICS}, volume 156 of {\em Advances in Imaging and Electron Physics}, pages
  1 -- 54. Elsevier.

\bibitem[Argyriou and Vlachos, 2003]{1224852}
Argyriou, V. and Vlachos, T. (2003).
\newblock Sub-pixel motion estimation using gradient cross-correlation.
\newblock In {\em Seventh International Symposium on Signal Processing and Its
  Applications, 2003. Proceedings.}, volume~2, pages 215--218 vol.2.

\bibitem[Argyriou and Vlachos, 2005]{1408930}
Argyriou, V. and Vlachos, T. (2005).
\newblock Performance study of gradient correlation for sub-pixel motion
  estimation in the frequency domain.
\newblock {\em IEE Proceedings - Vision, Image and Signal Processing},
  152(1):107--114.

\bibitem[Baltrusaitis et~al., 2016]{Baltru2016}
Baltrusaitis, T., Robinson, P., and Morency, L.~P. (2016).
\newblock Openface: an open source facial behavior analysis toolkit.
\newblock {\em IEEE Winter Conference on Applications of Computer Vision
  (WACV)}, pages 1--10.

\bibitem[Bettadapura, 2012]{Bettadapura2012}
Bettadapura, V. (2012).
\newblock Face expression recognition and analysis: the state of the art.
\newblock {\em Tech Report arXiv:1203.6722}, pages 1--27.

\bibitem[Bowers and Streinu, 2015]{Bowers2015}
Bowers, J.~C. and Streinu, I. (2015).
\newblock Lang’s universal molecule algorithm.
\newblock {\em Annals of Mathematics and Artificial Intelligence},
  74(3--4):371--400.

\bibitem[Buciu and Pitas, 2003]{Buciu2003}
Buciu, I. and Pitas, I. (2003).
\newblock Ica and gabor representation for facial expression recognition.
\newblock {\em ICIP}, 2:II--855.

\bibitem[Chowdhuri and Bojewar, 2016]{Chowdhuri2016}
Chowdhuri, M. A.~D. and Bojewar, S. (2016).
\newblock Emotion detection analysis through tone of user: A survey.
\newblock {\em International Journal of Advanced Research in Computer and
  Communication Engineering}, 5(5):859--861.

\bibitem[Chu et~al., 2017]{Chu2017}
Chu, W., la~Torre, F.~D., and Cohn, J. (2017).
\newblock Learning spatial and temporal cues for multi-label facial action unit
  detection.
\newblock {\em Autom Face and Gesture Conf}, 4.

\bibitem[Corneanu et~al., 2016]{Corneanu2016}
Corneanu, C., Simon, M., Cohn, J., and Guerrero, S. (2016).
\newblock Survey on rgb, 3d, thermal, and multimodal approaches for facial
  expression recognition: History, trends, and affect-related applications.
\newblock {\em IEEE transactions on pattern analysis and machine intelligence},
  38(8):1548--1568.

\bibitem[den Stock et~al., 2015]{Stock2015}
den Stock, J.~V., Winter, F.~D., de~Gelder, B., Rangarajan, J., Cypers, G.,
  Maes, F., Sunaert, S., and et. al. (2015).
\newblock Impaired recognition of body expressions in the behavioral variant of
  frontotemporal dementia.
\newblock {\em Neuropsychologia}, 75:496--504.

\bibitem[Ekman and Friesen, 1978]{Ekman1978}
Ekman, P. and Friesen, W. (1978).
\newblock The facial action coding system: A technique for the measurement of
  facial movement.
\newblock {\em Consulting Psychologists}.

\bibitem[Gudi et~al., 2015]{Gudi2015}
Gudi, A., Tasli, H., den Uyl, T., and Maroulis, A. (2015).
\newblock Deep learning based facs action unit occurrence and intensity
  estimation.
\newblock {\em In Automatic Face and Gesture Recognition (FG), 2015 11th IEEE
  International Conference and Workshops on}, 6:1--5.

\bibitem[Izard, 2013]{Izard2013}
Izard, C.~E. (2013).
\newblock Human emotions.
\newblock {\em Springer Science and Business Media}.

\bibitem[Kanade et~al., 2000]{Kanade2000}
Kanade, T., Cohn, J.~F., and Tian, Y. (2000).
\newblock Comprehensive database for facial expression analysis.
\newblock {\em Fourth IEEE International Conference on Automatic Face and
  Gesture Recognition}, pages 46--53.

\bibitem[Kumar and Chellappa, 2018]{Kumar2018}
Kumar, A. and Chellappa, R. (2018).
\newblock Disentangling 3d pose in a dendritic cnn for unconstrained 2d face
  alignment.

\bibitem[Lang, 1996]{Lang1996}
Lang, R.~J. (1996).
\newblock A computational algorithm for origami design.
\newblock {\em In Proceedings of the twelfth annual symposium on Computational
  geometry}, pages 98--105.

\bibitem[Li et~al., 2018]{Li2018}
Li, L., Zheng, W., Zhang, Z., Huang, Y., and Wang, L. (2018).
\newblock Skeleton-based relational modeling for action recognition.
\newblock {\em CoRR}, abs/1805.02556.

\bibitem[Li et~al., 2011]{Li2011}
Li, X., Adali, T., and Anderson, M. (2011).
\newblock Noncircular principal component analysis and its application to model
  selection.
\newblock {\em IEEE Transactions on Signal Processing}, 59(10):4516--4528.

\bibitem[Lokannavar et~al., 2015a]{Lokannavar2015}
Lokannavar, S., Lahane, P., Gangurde, A., and Chidre, P. (2015a).
\newblock Emotion recognition using eeg signals.
\newblock {\em International Journal of Advanced Research in Computer and
  Communication Engineering}, 4(5):54--56.

\bibitem[Lokannavar et~al., 2015b]{Lokannavar15}
Lokannavar, S., Lahane, P., Gangurde, A., and Chidre, P. (2015b).
\newblock Emotion recognition using eeg signals.
\newblock {\em Emotion}, 4(5):54--56.

\bibitem[Lucey et~al., 2010]{Lucey2010}
Lucey, P., Cohn, J., Kanade, T., Saragih, J., Ambadar, Z., and Matthews, I.
  (2010).
\newblock The extended cohn-kanade dataset (ck+): A complete dataset for action
  unit and emotion-specified expression.
\newblock {\em In Computer Vision and Pattern Recognition Workshops (CVPRW),
  2010 IEEE Computer Society Conference on}, pages 94--101.

\bibitem[McKeown et~al., 2012]{McKeown2012}
McKeown, G., Valstar, M., Cowie, R., Pantic, M., and Schroder, M. (2012).
\newblock The semaine database: Annotated multimodal records of emotionally
  colored conversations between a person and a limited agent.
\newblock {\em IEEE Transactions on Affective Computing}, 3(1):5--17.

\bibitem[Michel and Kaliouby, 2003]{Michel2003}
Michel, P. and Kaliouby, R.~E. (2003).
\newblock Real time facial expression recognition in video using support vector
  machines.
\newblock {\em In Proceedings of the 5th international conference on Multimodal
  interfaces}, pages 258--264.

\bibitem[Montenegro et~al., 2016]{Montenegro2016}
Montenegro, J., Gkelias, A., and Argyriou, V. (2016).
\newblock Emotion understanding using multimodal information based on
  autobiographical memories for alzheimer’s patients.
\newblock {\em ACCVW}, pages 252--268.

\bibitem[Montenegro and Argyriou, 2017]{FERNANDEZMONTENEGRO201742}
Montenegro, J. M.~F. and Argyriou, V. (2017).
\newblock Cognitive evaluation for the diagnosis of alzheimer's disease based
  on turing test and virtual environments.
\newblock {\em Physiology and Behavior}, 173:42 -- 51.

\bibitem[Ngo et~al., 2016]{LeNgo2016}
Ngo, A.~L., Oh, Y., Phan, R., and See, J. (2016).
\newblock Eulerian emotion magnification for subtle expression recognition.
\newblock {\em ICASSP}, pages 1243--1247.

\bibitem[Nicolaou et~al., 2011]{Nicolaou2011}
Nicolaou, M.~A., Gunes, H., and Pantic, M. (2011).
\newblock Continuous prediction of spontaneous affect from multiple cues and
  modalities in valence-arousal space.
\newblock {\em IEEE Transactions on Affective Computing}, 2(2):92--105.

\bibitem[Nicolle et~al., 2012]{Nicolle2012}
Nicolle, J., Rapp, V., Bailly, K., Prevost, L., and Chetouani, M. (2012).
\newblock Robust continuous prediction of human emotions using multiscale
  dynamic cues.
\newblock {\em 14th ACM conf on Multimodal interaction}, pages 501--508.

\bibitem[Pantic et~al., 2005]{Pantic2005}
Pantic, M., Valstar, M., Rademaker, R., and Maat, L. (2005).
\newblock Web-based database for facial expression analysis.
\newblock {\em IEEE international conference on multimedia and Expo}, pages
  317--321.

\bibitem[Park et~al., 2015]{Park2015}
Park, S., Lee, S., and Ro, Y. (2015).
\newblock Subtle facial expression recognition using adaptive magnification of
  discriminative facial motion.
\newblock {\em 23rd ACM international conference on Multimedia}, pages
  911--914.

\bibitem[Sariyanidi et~al., 2013]{Sariyanidi2013}
Sariyanidi, E., Gunes, H., Gkmen, M., and Cavallaro, A. (2013).
\newblock Local zernike moment representation for facial affect recognition.
\newblock {\em British Machine Vision Conf}.

\bibitem[Soleymani et~al., 2016]{Soleymani2016}
Soleymani, M., Asghari-Esfeden, S., Fu, Y., and Pantic, M. (2016).
\newblock Analysis of eeg signals and facial expressions for continuous emotion
  detection.
\newblock {\em IEEE Transactions on Affective Computing}, 7(1):17--28.

\bibitem[Soleymani et~al., 2012]{Soleymani2012}
Soleymani, M., Lichtenauer, J., Pun, T., and Pantic, M. (2012).
\newblock A multimodal database for affect recognition and implicit tagging.
\newblock {\em IEEE Transactions on Affective Computing}, 3(1):42--55.

\bibitem[Song et~al., 2015]{Song2015}
Song, Y., McDuff, D., Vasisht, D., and Kapoor, A. (2015).
\newblock Exploiting sparsity and co-occurrence structure for action unit
  recognition.
\newblock {\em In Automatic Face and Gesture Recognition (FG), 2015 11th IEEE
  International Conference and Workshops on}, 1:1--8.

\bibitem[Valstar et~al., 2017]{FERA2017}
Valstar, M., Sánchez-Lozano, E., Cohn, J., Jeni, L., Girard, J., Zhang, Z.,
  Yin, L., and Pantic, M. (2017).
\newblock Fera 2017: Addressing head pose in the third facial expression
  recognition and analysis challenge.
\newblock {\em arXiv preprint arXiv:1702.04174}, 19(1-–12):888--896.

\bibitem[Wadhwa et~al., 2016]{Wadhwa2016}
Wadhwa, N., Wu, H., Davis, A., Rubinstein, M., Shih, E., Mysore, G., Chen, J.,
  Buyukozturk, O., Guttag, J., Freeman, W., and Durand, F. (2016).
\newblock Eulerian video magnification and analysis.
\newblock {\em Communications of the ACM}, 60(1):87--95.

\bibitem[Weninger et~al., 2015]{Weninger2015}
Weninger, F., Wllmer, M., and Schuller, B. (2015).
\newblock Emotion recognition in naturalistic speech and language survey.
\newblock {\em Emotion Recognition: A Pattern Analysis Approach}, pages
  237--267.

\bibitem[Wu et~al., 2012]{Wu2012}
Wu, H., Rubinstein, M., Shih, E., Guttag, J., Durand, F., and Freeman, W.
  (2012).
\newblock Eulerian video magnification for revealing subtle changes in the
  world.
\newblock {\em ACM Transactions on Graphics}, 31:1--8.

\bibitem[Yan, 2017]{Yan2017}
Yan, H. (2017).
\newblock Collaborative discriminative multi-metric learning for facial
  expression recognition in video.
\newblock {\em Pattern Recognition}.

\bibitem[Yan et~al., 2018]{Yan2018}
Yan, S., Xiong, Y., and Lin, D. (2018).
\newblock Spatial temporal graph convolutional networks for skeleton-based
  action recognition.
\newblock {\em CoRR}, abs/1801.07455.

\bibitem[Yan et~al., 2014]{Yan2014}
Yan, W.~J., Li, X., Wang, S.~J., Zhao, G., Liu, Y.~J., Chen, Y.~H., and Fu, X.
  (2014).
\newblock Casme ii: An improved spontaneous micro-expression database and the
  baseline evaluation.
\newblock {\em PloS one}, 9(1):e86041.

\bibitem[Yi et~al., 2013]{Yi2013}
Yi, J., Mao, X., Xue, Y., and Compare, A. (2013).
\newblock Facial expression recognition based on t-sne and adaboostm2.
\newblock {\em GreenCom}, pages 1744--1749.

\bibitem[Zeng et~al., 2009]{Zeng2009}
Zeng, Z., Pantic, M., Roisman, G.~I., and Huang, T.~S. (2009).
\newblock A survey of affect recognition methods: Audio, visual, and
  spontaneous expressions.
\newblock {\em PAMI}, 31(1):39--58.

\bibitem[Zhao and Pietikinen, 2009]{Zhao2009}
Zhao, G. and Pietikinen, M. (2009).
\newblock Boosted multi-resolution spatiotemporal descriptors for facial
  expression recognition.
\newblock {\em Pattern recognition letters}, 30(12):1117--1127.

\end{thebibliography}

\end{document}